\documentclass{article}
\usepackage{spconf,amsmath,graphicx}
\usepackage{algorithm, algpseudocode}
\usepackage{hyperref}
\usepackage{amssymb}
\usepackage{amsmath}
\usepackage{graphicx}
\usepackage{subcaption,siunitx,booktabs}
\hypersetup{
    colorlinks=true,
    linkcolor=blue,
    filecolor=magenta,
    urlcolor=cyan,
    pdftitle={Sharelatex Example},
    bookmarks=true,
    pdfpagemode=FullScreen,
}

\usepackage{bm}

\def\vx{\mathbf{x}}
\def\vw{\mathbf{w}}
\def\vv{\mathbf{v}}
\def\vy{\mathbf{y}}
\def\vz{\mathbf{z}}
\def\vu{\mathbf{u}}
\def\mX{\mathbf{X}}
\def\mI{\mathbf{I}}
\def\mA{\mathbf{A}}
\def\samplingset{\mathcal{M}}
\def\mD{\mathbf{D}}
\def\cluster{\mathcal{C}}
\def\mQ{\mathbf{Q}}

\newcommand{\datasize}[1]{m_{#1}}
\def\featurelen{n}
\def\mT{\mathbf{T}}

\def\R{\mathbb{R}}
\DeclareMathOperator*{\argmin}{argmin}
\def\defeq{:=}
\newcommand{\nodes}{\mathcal{V}}
\newcommand{\edges}{\mathcal{E}}
\title{Federated Learning from Big Data over Networks}
\name{Y. Sarcheshmehpour, M. Leinonen and  A. Jung \thanks{This work has been partially supported by Academy of Finland (grant with decision nr. 331197).}}
\address{Aalto University}
\begin{document}
%
\maketitle
\begin{abstract}
This paper formulates and studies a novel algorithm for federated learning
from large collections of local datasets. This algorithm capitalizes on an intrinsic
network structure that relates the local datasets via an undirected ``empirical'' graph.
We model such big data over networks using a networked linear regression model.
Each local dataset has individual regression weights. The weights of close-knit
sub-collections of local datasets are enforced to deviate only little. This lends
naturally to a network Lasso problem which we solve using a primal-dual method.
We obtain a distributed federated learning algorithm via a message passing
implementation of this primal-dual method. We provide a detailed analysis of
the statistical and computational properties of the resulting federated learning algorithm.
\end{abstract}
\begin{keywords}
machine learning, federated learning, convex optimization, estimation, complex networks
\end{keywords}
\section{Introduction}
\label{sec:intro}
Federated learning is a recent paradigm for training machine learning models
in a collaborative fashion using distributed local datasets \cite{McMahan2017,Kairouz2019,LiTalwalkar2020}.
These methods are appealing for sensitive applications, such as healthcare, as they
do not require local raw data to be revealed to others \cite{AgarwalcpSGD2018}.
In contrast to existing federated learning techniques, we study a novel federated
learning method that leverages a well-defined network structure relating local datasets.

Network structures between local datasets can arise from spatio-temporal proximity
or statistical dependencies. The network structure allows to learn tailored models
for coherent subsets, or clusters, of local datasets instead of a single global model
which is delivered by most existing federated learning methods \cite{McMahan2017}.
As a case in point, consider the high-precision management of pandemics. Local datasets
are generated by smartphones and wearables of individuals \cite{Sigg2014}. These local
datasets are related via different network structures such as physical contact networks,
social networks \cite{NewmannBook}, and also Co-morbidity networks \cite{NetMedNat2010}.

This paper represents networked data conveniently using an undirected “empirical”
or “similarity” graph. Each node of this graph represents a local dataset which is
constituted by feature vectors and labels. We model the relation between features
and labels of a local dataset using a local Lasso problem (sparse linear regression).

The local regression models are coupled by requiring their parameter or weight vector
to have a small total variation. This essentially requires weight vectors to be approximately
constant for all local datasets within the same well-connected subset or cluster.
We frame federated learning as a convex network Lasso optimization problem
which we solve using an efficient and robust primal-dual method \cite{ProximalMethods,NetworkLasso,LocalizedLinReg2019}.

This paper significantly extends our previous work on localized linear
regression and classification \cite{LocalizedLinReg2019,Tran2020,LogisticNLassoAsilomar2018}.
In particular, we allow local datasets to contain many individual data points.
Moreover, our method can be used with non-linear local models such as
Lasso and its generalizations \cite{HastieWainwrightBook}.

{\bf Notation.} 
The identity matrix of size $n\!\times\!n$ is denoted $\mathbf{I}_{n}$ with the subscript omitted
if the size $n$ is clear from context.
 The Euclidean norm of a vector $\vw\!=\!(w_{1},\ldots,w_{\featurelen})^{T}$ is $\| \vw \|_{2} \!\defeq\!\sqrt{\sum_{r=1}^{\featurelen} w_{r}^{2}}$ and the $\ell_{1}$ norm 
$\| \vw \|_{1} \!\defeq\!\sum_{r=1}^{\featurelen} |w_{r}|$. It will be convenient to 
use the notation $(1/2\tau)$ instead of $(1/(2\tau))$. 
We will need the scalar clipping function $\mathcal{T}^{(\lambda)}(w)\!\defeq\!\lambda w/|w|$ 
for $| w |\!\geq\!\lambda$ and $\mathcal{T}^{(\lambda)}(w) \defeq w$ otherwise. 





\section{PROBLEM FORMULATION}
\label{sec:format}
\begin{figure}[htb]
\centering
\includegraphics[scale=0.2]{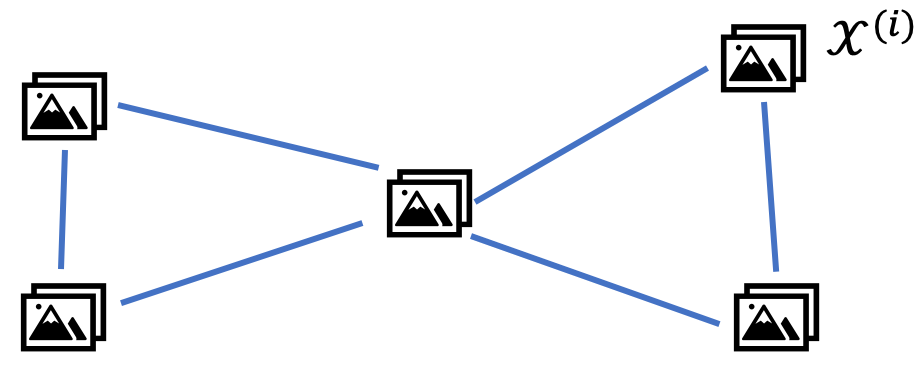}
\caption{Empirical graph of local datasets $\mathcal{X}^{(i)}$.}
\label{fig_local_dataset}
\end{figure}

We model local datasets by an undirected ``empirical'' graph
$\mathcal{G}=(\nodes,\edges,\mA)$ (see Figure \ref{fig_local_dataset}).

Each node $i\!\in\!\nodes$ represents a local dataset
$$ \mathcal{X}^{(i)} \defeq \big(\vx^{(i,1)},y_{1}^{(i)}\big),\ldots,\big(\vx^{(i,\datasize{i})},y_{\datasize{i}}^{(i)}\big).$$
We assume that features $\vx^{(i,j)}$ of any data point can be determined easily.
In contrast, the acquisition of labels $y_{j}^{(i)}$ is costly (requiring human expert labour).
Thus, we assume to have access only to the labels of a small training set
\begin{equation}
\mathcal{M} = \{i_1,...,i_M\} \subseteq \nodes.
\end{equation}
The undirected edges $\{i,j\}\!\in\!\edges$ of the empirical graph connect
similar datasets $\mathcal{X}^{(i)}$, $\mathcal{X}^{(j)}$ with the  strength
of the similarity quantified by the entry $A_{ij}\!>\!0$. The neighbourhood of
the node $i \in \nodes$ is denoted $\mathcal{N}_{i} \defeq \{ j \in \nodes: \{i,j\} \in \edges\}$.

We aim at learning a local (node-wise) predictor $h(\vx;\vw^{(i)})$ for each node 
$i\!\in\!\nodes$. The local predictors are parametrized by the weight vectors $\vw^{(i)}$. 
For numeric labels $y \in \mathbb{R}$, we use the output $h(\vx;\vw^{(i)})$ directly 
as the predicted label $\hat{y}=h(\vx;\vw^{(i)})$. For binary labels $y \in \{0,1\}$ we 
classify $\hat{y} = 1$ for $h(\vx;\vw^{(i)}) \geq 0$ and $\hat{y} = 0$ otherwise. 

Our approach to learning the weights $\mathbf{w}^{(i)}$ exploits the intrinsic network structure
relating the local datasets. We use a clustering assumption of having similar statistical properties
of local datasets that form tight-knit subsets or clusters. We show theoretically
and empirically that enforcing the learnt weight vectors $\vw^{(i)}$ to be similar for
well-connected datasets allows to share statistical strength between local datasets.


We interpret the weight vectors $\vw^{(i)}$ as the values of a graph signal 
$\vw\!:\!\nodes\!\rightarrow\!\R^{\featurelen}$ which assigns node $i\!\in\!\nodes$ 
the vector $\vw^{(i)}\!\in\!\R^{\featurelen}$. With a slight abuse of notation, 
we denote the set of all vector-valued node signals as $\mathbb{R}^{\featurelen|\nodes|} \defeq \{ \vw: \nodes  \rightarrow \R^{\featurelen} : i \mapsto \vw^{(i)} \}$. Each graph signal $\vw\!\in\!\mathbb{R}^{\featurelen|\nodes|}$ represents 
networked predictor, parametrized by $\vw^{(i)} \in \mathbb{R}^{\featurelen}$ for $i \in \nodes$.  

A good predictor should have small empirical loss
\begin{equation} \label{eq:4}
 \widehat{E}(\vw) \defeq \sum_{i \in \mathcal{M}} \mathcal{L}\big(\mathcal{X}^{(i)},\vw^{(i)} \big). 
\end{equation}
The loss function $\mathcal{L}\big(\mathcal{X}^{(i)},\vw^{(i)} \big)$ measures the quality
of applying the predictor $h^{(i)}(\vx) \defeq \vx^{T} \vw^{(i)}$ to the local dataset $\mathbf{X}^{(i)}$
and $\vy^{(i)}$. Section \ref{sec_federatedml} discusses three choices for the loss function.

\section{NETWORK LASSO}
\label{sec_nLasso}

The criterion \eqref{eq:4} by itself is not enough for guiding the learning of a predictor $\vw$ 
since \eqref{eq:4} completely ignores the weights $\vw^{(i)}$ at unlabeled nodes 
$i \in \mathcal{V} \backslash \mathcal{M}$. We need to impose some additional 
structure on the predictor $\vw$. To this end, we require the predictor $\vw$ 
to conform with the cluster structure of the empirical graph $\mathcal{G}$. 

To obtain similar predictors $\vw^{(i)}\!\approx\!\vw^{(j)}$ for nodes $i,j\in \nodes$ 
belonging to the same cluster, we enforce a small total variation (TV)
\begin{equation} \label{eq:5}
\| \vw \|_{\rm TV}  \defeq \sum_{i, j \in \mathcal{E}} A_{ij} ||\vw^{(j)} - \vw^{(i)}||_1.
\end{equation}
Minimizing TV forces weights $\vw^{(i)}$ to only change over few edges with relatively 
small weights $A_{i,j}$. The predictor optimally balancing between empirical 
error \eqref{eq:4} with TV is
\begin{equation} \label{eq:6}
\hat{\vw} \in \underset{{\bf w} \in \mathbb{R}^{\featurelen|\nodes|}}{\mathrm{arg \ min}}\ \hat{E}(\vw) + \lambda \|\vw\|_{\rm TV}.
\end{equation}
Note that \eqref{eq:6} does not enforce the predictions $\hat{y}^{(i)}$
themselves to be clustered, but the weight vectors $\hat{\vw}^{(i)}$ of predictors.

The convex optimization problem \eqref{eq:6} is a special case of nLasso \cite{NetworkLasso}.
The parameter $\lambda > 0$ in \eqref{eq:4} allows to trade small TV $||\hat{\bf w}||_{\rm TV}$
against small error $\hat{E}(\hat{\vw})$ \eqref{eq:4}. The choice of $\lambda$ can be guided
by cross validation \cite{hastie01statisticallearning}. 

Let us define the block incidence matrix
$\mD \in \mathbb{R}^{\featurelen |\edges| \times \featurelen |\nodes|}$ as follows:
$D_{e, i} = {\bf I}$ for $e\!=\!\{i, j\} \in \edges$ with some  $j\!>\!i$,  $D_{e, i} = - {\bf I}$
for $e\!=\!\{i, j\}$ with some  $i\!>\!j$ and $D_{e, i} = \mathbf{0}$ otherwise.
Then we can reformulate \eqref{eq:6} as
\begin{equation} \label{equ_nLasso}
\hat{\vw} \in \underset{{\vw} \in \mathbb{R}^{\featurelen|\nodes|}}{\mathrm{arg \ min}}\ f(\vw)\!+\!g(\mD\vw)
\end{equation}
with
\begin{align}
\label{equ_components_nLasso}
f(\vw) &\!\defeq\!\sum_{i \in \mathcal{M}} \mathcal{L}\big(\mathcal{X}^{(i)}, \vw^{(i)} \big) \mbox{, and } g(\vu) \!\defeq\!\lambda \sum_{e \in \edges} A_{e} \big\| \vu^{(e)} \big\|_1 \nonumber  \\
\mbox{ using }  \vu & = \left ( (\vu^{(1)})^T ,\ldots, (\vu^{(\edges)})^T \right)^T \in  \R^{\featurelen|\edges|}.
\end{align}

\section{Federated Learning Algorithm}
\label{sec_federatedml}

We solve \eqref{equ_nLasso} jointly with the dual problem
\begin{equation}
\label{equ_dual_nLasso}
\max_{\vu \in \mathbb{R}^{\featurelen|\edges|}} -g^{*}(\vu) - f^{*}(-\mD^{T} \vu).
\end{equation}
The problem \eqref{equ_dual_nLasso} optimizes a vector-valued signal
$\vu\!\in\!\mathbb{R}^{\featurelen|\edges|}$ which maps each edge $e\!\in\!\edges$ to
to some vector $\vu^{(e)}\!\in\!\mathbb{R}^{\featurelen}$. The objective function 
\eqref{equ_dual_nLasso} is composed of the convex conjugates
\begin{align}
\label{equ_conv_conjugate_g_dual_proof}
g^{*}(\vu) &\defeq \sup_{\vz \in \mathbb{R}^{\edges}} \vu^{T}\vz - g(\vz)  \nonumber \\
& \stackrel{\eqref{equ_components_nLasso}}{=} \sup_{\vz \in \mathbb{R}^{\edges}} \vu^{T}\vz - \lambda \sum_{e \in \edges} A_{e} \|\vz^{(e)}\|_{1} \nonumber\\
&= \begin{cases}
\infty &\text{if $|u_{j}^{(e)}| > \lambda A_e \mbox{ for some } e \in \edges$}\\
0 &\text{otherwise,}
\end{cases}
\end{align}
and
\begin{align}
\label{equ_dual_f_fun}
f^{*}(\vw) \!\defeq\! \sup_{\vz \in \mathbb{R}^{\featurelen|\nodes|}} \vw^{T}\vz- f(\vz).
\end{align}

The duality between \eqref{equ_nLasso} and \eqref{equ_dual_nLasso} is made precise
in \cite[Ch.\ 31]{RockafellarBook} (see also \cite[Sec. 3.5]{pock_chambolle_2016}).
The optimal values of both problems are the same \cite[Cor.\ 31.2.1]{RockafellarBook},
\begin{equation}
\label{equ_equal_primal_dual}
\min_{\vw \in \mathbb{R}^{\featurelen |\nodes|}} \hspace*{-1mm} f(\vw)\!+\!g(\mD \vw ) \!=\! \hspace*{-1mm}\max_{\vu \in \mathbb{R}^{\featurelen|\edges|}} -g^{*}(\vu)\!-\!f^{*}(-\mD^{T} \vu).
\end{equation}
A necessary and sufficient condition for $\hat{\vw}$ to solve \eqref{equ_nLasso}
and $\hat{\vu}$ to solve \eqref{equ_dual_nLasso} is
\begin{equation}
\label{equ_opt_condition_Rocka_KKT}
-\mD^{T} \hat{\vu} \in \partial f(\hat{\vw}) \mbox{ , and } \mD \hat{\vw} \in  \partial  g^{*}(\hat{\vu}).
\end{equation}
The coupled conditions \eqref{equ_opt_condition_Rocka_KKT} are, in turn, equivalent to
\begin{align} \label{eq:13}
\hat{\vw} -\mT \mD^T \hat{\vu} & \in ({\bf I} + \mT \partial f)(\hat{\vw}), \nonumber \\
\hat{\vu} + {\bf \Sigma} \mD\hat{\vw} & \in ({\bf I} + {\bf \Sigma} \partial g^*)(\hat{\vu}).
\end{align}
The positive semi-definite block-diagonal matrices 
\begin{align} \label{eq:14}
\big( {\bf \Sigma} \big)_{e,e}= \sigma^{(e)} {\bf I}_n, \mbox{ for } e\!\in\!\edges  \mbox{, }
\big(\mT\big)_{i,i}  = \tau^{(i)} \mI \mbox{ for } i\!\in\!\nodes, 
\end{align}
with $\sigma^{(e)}\!=\!1/2$ for $e\!\in\!\edges$ and $\tau^{(i)}\!=\!1/ | \mathcal{N}_{i}|$ for $i\!\in\!\nodes$, 
serve as a pre-conditioning to speed up the convergence of the resulting iterative algorithm 
(see \cite{PrecPockChambolle2011}). 

The coupled conditions \eqref{eq:13} represent a fixed-point characterization
of the solutions to nLasso \eqref{equ_nLasso} and its dual \eqref{equ_dual_nLasso}.
We solve the optimality condition \eqref{eq:13} iteratively using the following
fixed-point iterations (see \cite{pock_chambolle_2016})
\begin{align}
\vw_{k+1} &\!\defeq\! \big( \mI \!+\! \mT \partial f \big)^{-1} (\vw_k - \mT \mD^T \vu_k) \label{equ_pd_updates_resolvents_primal} \\
\vu_{k+1} &\!\defeq\! \big( \mI \!+\! {\bf \Sigma} \partial g^* \big)^{-1} ({\bf u}_k\!+\!{\bf \Sigma}  \mD ( 2\vw_{(k+1)}\!-\!\vw_{(k)})).   \label{equ_pd_updates_resolvents_dual}
\end{align}
The updates \eqref{equ_pd_updates_resolvents_primal} can be evaluated by
using the identity \cite{PrecPockChambolle2011}
\begin{align}
\big(\mI\!+\!\mT \partial f \big)^{-1} (\vw) 
 \!=\! \argmin_{\vv \in \mathbb{R}^{\featurelen|\nodes|}} f(\vv)&\!+\!(1/2) \|\vv\!-\!\vw\|^{2}_{\mT^{-1}} \label{equ_opt_form_resolvent}
\end{align}
and a similar identity for $\big( \mI \!+\! {\bf \Sigma} \partial g^* \big)^{-1}$.

Inserting \eqref{equ_components_nLasso} into \eqref{equ_opt_form_resolvent} reveals that the primal update
decomposes into separate updates at each node $i \in \nodes$,
\begin{align}
\label{equ_node_wise_primal_update}
\vw^{(i)}_{k+1} & \!=\! \mathcal{PU}_{i} \big\{ \vw^{(i)}_k -  \tau^{(i)} \sum_{e \in \edges} D_{e,i} \vu^{(e)}_k \big \}
\end{align}
with the primal update operator
\begin{equation}
\label{equ_node_wise_primal_update_min}
 \mathcal{PU}_{i} \big\{ \vv \big \}\defeq \argmin_{\vz \in \mathbb{R}^{\featurelen}} \mathcal{L}(\mathcal{X}^{(i)},\vz)\!+\!(1/2\tau_{i}) \|\vv\!-\!\vz\|^{2}. 
\end{equation}
The operator \eqref{equ_node_wise_primal_update_min} and associated 
node-wise primal update and \eqref{equ_node_wise_primal_update} depend 
(via \eqref{equ_components_nLasso}) on the choice for the loss function $\mathcal{L}$.

For convex loss functions $\mathcal{L}(\mathcal{X}^{(i)},\vz)$ (see \eqref{eq:4}),
the sequences ${\bf w}_{k}$ and ${\bf u}_{k}$ obtained from iterating \eqref{equ_pd_updates_resolvents_primal},
\eqref{equ_pd_updates_resolvents_dual} converge to solutions of \eqref{equ_nLasso}
and \eqref{equ_dual_nLasso}. This convergence is guaranteed for any initialization ${\bf w}_{0}$ and ${\bf u}_{0}$ \cite{PrecPockChambolle2011}.

Another appealing property of the updates \eqref{equ_pd_updates_resolvents_primal},
\eqref{equ_pd_updates_resolvents_dual} is that they are robust against errors. This
property is important for applications where the resolvent operator \eqref{equ_opt_form_resolvent}
can be evaluated approximately only. An important example when this happens is
obtained when using the logistic loss function \eqref{equ_def_logistic_loss}
(see Section \ref{sec_log_reg}).

We summarize the primal-dual method for solving nLasso \eqref{equ_nLasso} and
its dual \eqref{equ_dual_nLasso} in Algorithm \ref{alg1}. This algorithm is to be
understood as a template for specific learning algorithms that are obtained by
evaluating \eqref{equ_node_wise_primal_update} for given choice for the loss
function in \eqref{eq:4}.

\begin{algorithm}[htbp]
	\caption{Primal-Dual Method for Networked Federated Learning}
	\label{alg1}
	{\bf Input}: $\mathcal{G}\!=\!(\nodes, \edges,\mA),  \{ \mX^{(i)} \}_{i \in \nodes} ,  \mathcal{M}, \{ \mathcal{X}^{(i)} \}_{i \in \samplingset}, \lambda$\\
	{\bf Initialize}: $k\!\defeq\!0$;$\hat{\vw}_0\!\defeq\!{\bf 0}$;$\hat{\vu}_0 \!\defeq\! {\bf 0}$; ${\bf \Sigma}$ and $\mT$
	using \eqref{eq:14}; 
	\begin{algorithmic}[1]
		\While{stopping criterion is not satisfied}
		\For{labeled local datasets $ i \in \mathcal{M} $}
		\State $\hat{\vw}_{k+1}^{(i)} \defeq \mathcal{PU}_{i} \big\{ \hat{\vw}^{(i)}_k -  \tau^{(i)} \sum_{e \in \edges} D_{e,i} \hat{\vu}^{(e)}_k \big \}$ \label{primal_udpate}
		\EndFor
		\For{non-labeled local datasets $ i \in \nodes \setminus \mathcal{M} $}
		\State $\hat{\vw}_{k+1}^{(i)} \defeq \hat{\vw}^{(i)}_k -  \tau^{(i)} \sum_{e \in \edges} D_{e,i} \hat{\vu}^{(e)}_k $ \label{equ_non_labled_pu}
		\EndFor
		\State $\vu = \hat{\vu}_k + {\bf \Sigma} \mD ( 2\hat{\bf w}_{k+1} - \hat{\vw}_k) $  \label{equ_update_dual}
		\State $u_{j}^{(e)}\!\defeq\!\mathcal{T}^{(\lambda A_{e})} \big( u_{j}^{(e)} \big)$ for $e\in \edges, j \in \{1,\ldots,\featurelen\}$
		\State $\hat{\vu}_{k+1} \defeq \vu$
		\State $k\!\defeq\!k\!+\!1$
		\EndWhile
	\end{algorithmic}
\end{algorithm}

Algorithm \ref{alg1} can be implemented as a message passing on the empirical graph. 
The application of the block-incidence matrix $\mathbf{D}$ and its transpose $\mathbf{D}^{T}$ 
involves, for each node and edge,only neighbouring nodes and edges. Thus, Algorithm \ref{alg1} is 
scalable to massive collections of local datasets.

Algorithm \ref{alg1} combines the information contained in the local datasets
with their network structure to iteratively improve the weight vectors $\hat{\vw}^{(i)}_{k}$
for each node $i\in \nodes$. Step \ref{primal_udpate} adapts the current weight vectors 
$\hat{\vw}^{(i)}_{k}$ to better fit the labeled local datasets $\mathcal{X}^{(i)}$ for $i \in \mathcal{M}$.
These updates are then propagated to the weight vectors at unlabeled nodes
$i \in \nodes \setminus \mathcal{M}$ via steps \ref{equ_non_labled_pu} and \ref{equ_update_dual}.


\subsection{Federated Networked Linear Regression}
\label{sec_net_lin_reg}

We now discuss how Algorithm \ref{alg1} can be used to learn a node-wise linear predictor
\begin{equation}
\label{equ_def_lin_pred}
\hat{y} = h(\mathbf{x};\mathbf{w}^{(i)}) = \mathbf{x}^{T} \mathbf{w}^{(i)}.
\end{equation}
To measure how well a given network of node-wise linear predictor agrees
with labeled datasets $\mathcal{X}^{(i)}$, for $i \in \mathcal{M}$, we use the
squared error loss 
\begin{equation}
\label{equ_squared_error_loss}
 \hspace*{-2mm}\mathcal{L}\big(\mathcal{X}^{(i)}, \vv \big)\!=\!(1/\datasize{i}) \sum_{r=1}^{\datasize{i}} \big( y_{r}^{(i)}\!-\! \vv^{T} \vx^{(i,r)}  \big)^{2}
\end{equation}

Inserting \eqref{equ_squared_error_loss}
into \eqref{equ_node_wise_primal_update_min}, yields
\begin{equation}
 \mathcal{PU}_{i} \big( \vv \big)= \left [ \mI \!+\!2 \tau^{(i)} \tilde{\bf Q}^{(i)} \right ]^{-1} \left [\vv + 2 \tau^{(i)} \tilde{\vy}^{(i)} \right].
\end{equation}
Here, we used the squared feature matrix $$\mQ^{(i)}\!\defeq\!\big(\mX^{(i)}\big)^{T} \mX^{(i)} \mbox{ with } \mX^{(i)} \defeq \big(\vx^{(i,1)},\ldots,\vx^{(i,\datasize{i})}\big)^{T}$$
and the normalized label vector $$\tilde{\vy}^{(i)}\!\defeq\!\big(\mX^{(i)}\big)^{T} \vy^{(i)} \mbox{ with } \vy^{(i)} \defeq \big(y^{(i)}_{1},\ldots,y_{\datasize{i}}^{(i)}\big)^{T}.$$

\subsection{Federated Networked Lasso}
\label{sec_net_lin_reg}

Algorithm \ref{alg1} for networked linear regression (see Section \ref{sec_net_lin_reg})
can only be expected to work well if the number of data points in each local dataset
is larger than the number of features $\featurelen$.

Many application domains involve high-dimensional local datasets $\mathcal{X}^{(i)}$, where
the number of data points is much smaller than the length of feature vectors, $\datasize{i} \ll \featurelen$ \cite{BuhlGeerBook,HastieWainwrightBook}.
This high-dimensional regime requires some form of regularization for learning a linear predictor \eqref{equ_def_lin_pred}.
The Lasso is obtained from the regularized loss function \cite{HastieWainwrightBook}
$$\mathcal{L}\big(\mathcal{X}^{(i)},\vv \big)\defeq (1/\datasize{i}) \sum_{r=1}^{\datasize{i}} \big( \vv^{T} \vx^{(i,r)}\!-\!y_{r}^{(i)} \big)^2  \nonumber \\
+\lambda \| \vv \|_{1}.$$
Plugging this into \eqref{equ_node_wise_primal_update_min}, yields the primal update operator
\begin{align}
 \mathcal{PU}_{i} \big( \vw^{(i)} \big)& =\argmin_{\vv \in \mathbb{R}^{\featurelen}} (1/\datasize{i}) \sum_{r=1}^{\datasize{i}} \big( \vv^{T} \vx^{(i,r)}\!-\!y_{r}^{(i)} \big)^2  \nonumber \\
 & +\lambda \| \vv \|_{1}  +(1/2\tau_{i}) \big( \vv\!-\!\vw^{(i)} \big)^2.
\end{align}

\subsection{Federated Networked Logistic Regression}
\label{sec_log_reg}
We discuss how Algorithm \ref{alg1} can be used to learn a networked
linear classifier for binary labels $y \in \{0,1\}$.
As in Section \ref{sec_net_lin_reg}, we aim at learning a networked linear
predictor. In contrast to Section \ref{sec_net_lin_reg}, we quantize its output
to obtain the predicted label $\hat{y}\!=\!1$ if $h(\vx;\vw^{(i)})\!>\!0$ and $\hat{y}\!=\!0$ otherwise.

To measure the quality of a given linear classifier we use the logistic loss

\begin{align}
\label{equ_def_logistic_loss}
\mathcal{L}(\mathcal{X}^{(i)}, \vv)  & = (-1/m_i ) \sum_{r=1}^{m_i} \bigg[  y^{(i)}_r {\rm log} \left (\sigma({\bf v}^T {\bf x}^{(i, r)}) \right) \nonumber \\
+ & (1 - y^{(i)}_r) {\rm log} \left (1 - \sigma({\bf v}^T  {\bf x}^{(i, r)}) \right)   \bigg]. 
\end{align}

In general, there is no closed-form expression
for the update \eqref{equ_node_wise_primal_update} when using the logistic
loss \eqref{equ_def_logistic_loss}. However, the update \eqref{equ_def_logistic_loss}
amounts to an unconstrained minimization of a smooth convex function
\eqref{equ_node_wise_primal_update_min}. Such optimization problems can
be solved efficiently with established iterative algorithms \cite[Ch. 4]{hastie01statisticallearning}.

%
%

%
%
%

\section{Numerical Experiments}
\label{sec:majhead}

To empirically evaluate the accuracy of Algorithm \ref{alg1}, we
apply it to a synthetic dataset. We generate the empirical graph $\mathcal{G}$
using the stochastic block model with two clusters $|\cluster_1|=|\cluster_2|=150$ \cite{Mossel2012,AbbeSBM2018}. Two nodes within the same cluster are connected by an edge with probability $p_{\rm in}$, Two nodes from different clusters are connected by an edge with probability $p_{\rm out}$.


Each node $i\in \nodes$ represents a local dataset consisting of $5$ feature vectors
$\vx^{(i,1)},\ldots,\vx^{(i,5)} \in \mathbb{R}^{2}$. The feature vectors are i.i.d. realizations
of a standard Gaussian random vector $\vx \sim \mathcal{N}(\mathbf{0},\mathbf{I})$.
The labels $y_{1}^{(i)},\ldots,y_{5}^{(i)} \in \mathbb{R}$ of the nodes $i \in \mathcal{V}$ are generated according to
the linear model $y_{r}^{(i)} = \big( \mathbf{x}^{(i,r)} \big)^{T} \overline{\vw}^{(i)}$ with weight
vector $\overline{\vw}^{(i)} = \big(2,2\big)^{T}$ for $i \in \cluster_1$ and
$\overline{\vw}^{(i)} = \big(-2,2\big)^{T}$ for $i \in \cluster_2$.

\begin{figure}[!tbp]
  \centering
  \begin{minipage}[b]{0.23\textwidth}
    \includegraphics[width=\textwidth]{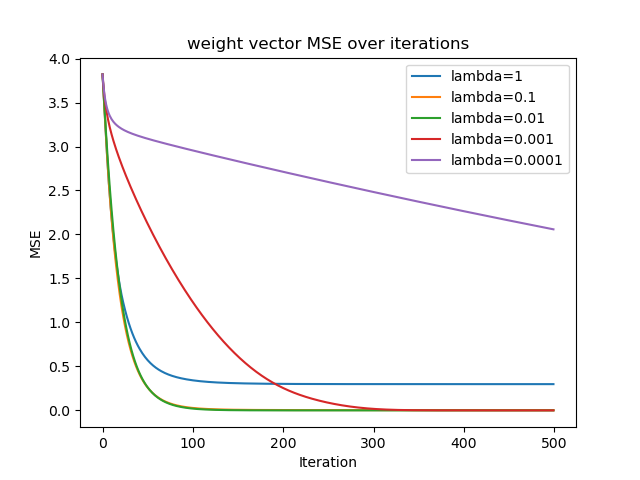}
	\caption{MSE incurred by the weights obtained after a varying number 
	of iterations used in Algorithm \ref{alg1}.}
	\label{fig:MSE_iteration}
  \end{minipage}
  \hfill
  \begin{minipage}[b]{0.23\textwidth}
    \includegraphics[width=\textwidth]{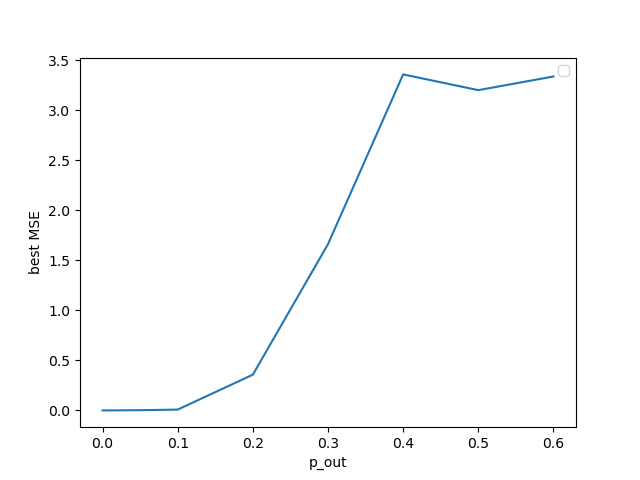}
\caption{MSE incurred by different $p_{out}$ for a fixed $p_{in}=\!1/2$. \\}
\label{fig:MSE_pout}
  \end{minipage}
\end{figure}

To learn the weight $\overline{\vw}^{(i)}$, we apply Algorithm \ref{alg1} to a training set $\samplingset$ 
obtained by randomly selecting $30$ nodes. We run Algorithm \ref{alg1} for different choices of $p_{\rm out}$ with a fixed $p_{\rm in}\!=\!1/2$ (fig ~\ref{fig:MSE_pout}).

By fixing $p_{\rm out}\!=\!10^{-3}$, we run Algorithm \ref{alg1} for different choices of $\lambda$ 
and a fixed number of $500$ iterations. We measure the quality of the learnt weight vectors $\hat{w}^{(i)}$ by 
the mean-squared error (MSE)  (fig ~\ref{fig:MSE_iteration})
\begin{equation}
\varepsilon(\hat{\vw}) \defeq (1/|\nodes|) \sum_{i \in \nodes \setminus \mathcal{M} } \big\| \overline{\vw}^{(i)} - \hat{\vw}^{(i)} \big\|^{2}_{2}.
\end{equation}
The tuning parameter $\lambda$ in \eqref{eq:6} is manually chosen, guided by the 
resulting MSE, as $\lambda=10^{-3}$. We compare the MSE of Algorithm \ref{alg1} with 
plain linear regression and decision tree regression in Table ~\ref{tab:sbm_tables}.




\begin{table}

   \begin{tabular}{c|c|c|c|c}
      \toprule
      Method Name  &  training MSE & test MSE \\
      \midrule
      our  method & 1.7e-6 & 1.8e-6 \\
      simple linear regression    &   4.04 & 4.51  \\
      decision tree regression   &   4.21 & 4.87  \\
      \bottomrule
   \end{tabular}

\caption{MSE achieved by Algorithm \ref{alg1} which leverages the network 
	structure encoded by the empirical graph $\mathcal{G}$. We also report 
	the MSE achieved by plain linear regression and decision tree regression applied 
	to the concatenation of all local datasets, ignoring the network structure. } \label{tab:sbm_tables}
\end{table}

\bibliography{/Users/alexanderjung/Literature}
\bibliographystyle{plain}

\end{document}